\documentclass{article}

\usepackage{arxiv}

\usepackage[english]{babel}

\usepackage{multicol}
\usepackage[font=small]{caption}
\usepackage{graphicx}
\usepackage[utf8]{inputenc} 
\usepackage[T1]{fontenc}    
\usepackage{hyperref}       
\usepackage{url}            
\usepackage{booktabs}       
\usepackage{amsfonts}       
\usepackage{nicefrac}       
\usepackage{microtype}      
\usepackage{lipsum}

\title{Short-Text Classification Using Unsupervised Keyword Expansion}

\author{
  Duncan Cameron-Steinke \thanks{Also affiliated to: Department of Physics, Engineering Physics and Astronomy, Queen's University Kingston, Canada}\\
  Institute for InfoComms Research\\
  A*STAR\\
  Singapore, Singapore \\
  \texttt{cameronsteinke.d@queensu.ca} \\
}

\begin{document}
\maketitle

\begin{abstract}
\label{abstract}

Short-text classification, like all data science, struggles to achieve high performance using limited data. As a solution, a short sentence may be expanded with new and relevant feature words to form an artificially enlarged dataset, and add new features to testing data. This paper applies a novel approach to text expansion by generating new words directly for each input sentence, thus requiring no additional datasets or previous training. In this unsupervised approach, new keywords are formed within the hidden states of a pre-trained language model and then used to create extended pseudo documents. The word generation process was assessed by examining how well the predicted words matched to topics of the input sentence. It was found that this method could produce 3-10 relevant new words for each target topic, while generating just 1 word related to each non-target topic. Generated words were then added to short news headlines to create extended pseudo headlines. Experimental results have shown that models trained using the pseudo headlines can improve classification accuracy when limiting the number of training examples.
\end{abstract}


\begin{multicols}{2}
\section{Introduction} 
\label{Introduction}

The web has provided researchers with vast amounts of unlabeled text data, and enabled the development of increasingly sophisticated language models which can achieve state of the art performance despite having no task specific training \cite{devlin2018bert, Radford, howard2018universal}. It is desirable to adapt these models for bespoke tasks such as short text classification. \newline
Short-text is nuanced, difficult to model statistically, and sparse in features, hindering traditional analysis \cite{7113309}. These difficulties become further compounded when training is limited, as is the case for many practical applications. \newline
This paper provides a method to expand short-text with additional keywords, generated using a pre-trained language model. The method takes advantage of general language understanding to suggest contextually relevant new words, without necessitating additional domain data. The method can form both derivatives of the input vocabulary, and entirely new words arising from contextualised word interactions and is ideally suited for applications where data volume is limited. 

\begin{Figure}
 \centering
 \includegraphics[width=\linewidth]{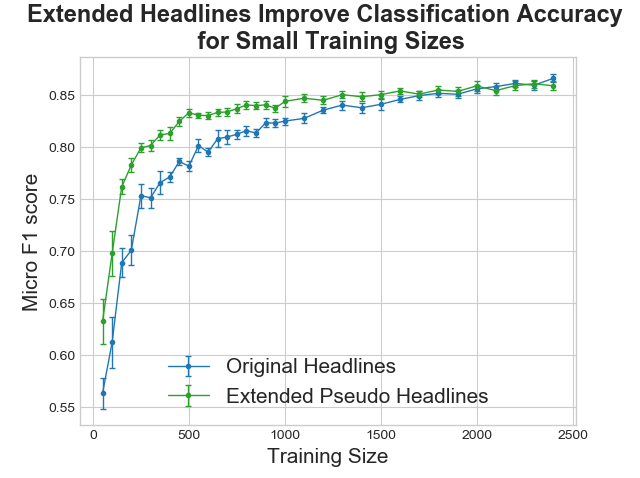}
 \captionof{figure}{Binary Classification of short headlines into 'WorldPost' or 'Crime' categories, shows improved performance with extended pseudo headlines when the training set is small.  Using: Random forest classifier, 1000 test examples, 10-fold cross validation.}
 \label{fig:Worldpost_vs_Crime}
\end{Figure}

\section{Literature Review}

Document expansion methods have typically focused on creating new features with the help of custom models. Word co-occurrence models \cite{bicalho2017general}, topic modeling \cite{zhang2016improving}, latent concept expansion \cite{Metzler2007}, and word embedding clustering \cite{Wang2016}, are all examples of document expansion methods that must first be trained using either the original dataset or an external dataset from within the same domain. The expansion models may therefore only be used when there is a sufficiently large training set. \newline

Transfer learning was developed as a method of reducing the need for training data by adapting models trained mostly from external data \cite{pan2010survey}. Transfer learning can be an effective method for short-text classification and requires little domain specific training data \cite{Phan2008, long2012tcsst}, however it demands training a new model for every new classification task and does not offer a general solution to sparse data enrichment. \newline

Recently, multi-task language models have been developed and trained using ultra-large online datasets without being confined to any narrow applications \cite{devlin2018bert, Radford, howard2018universal}. It is now possible to benefit from the information these models contain by adapting them to the task of text expansion and text classification. \newline

This paper is a novel approach which combines the advantages of document expansion, transfer learning, and multitask modeling. It expends documents with new and relevant keywords by using the BERT pre-trained learning model, thus taking advantage of transfer learning acquired during BERT's pretraining. It is also unsupervised and requires no task specific training, thus allowing the same model to be applied to many different tasks or domains.

\section{Procedures}
\label{Methods}
\subsection{Dataset}
\label{Dataset}
The News Category Dataset\footnote{Further information related to the dataset and its author can be found here: https://rishabhmisra.github.io/publications/} \cite{dataset} is a collection of headlines published by HuffPost \cite{Huffpost} between 2012 and 2018, and was obtained online from Kaggle \cite{RishabhMisra}. The full dataset contains 200k news headlines with category labels, publication dates, and short text descriptions. For this analysis, a sample of roughly 33k headlines spanning 23 categories was used. Further analysis can be found in table \ref{tab:data info} in the appendix.

\subsection{Word Generation}
\label{word generation}
Words were generated using the BERT pre-trained model developed and trained by Google AI Language \cite{devlin2018bert}. BERT creates contextualized word embedding by passing a list of word tokens through 12 hidden transformer layers and generating encoded word vectors. To generate extended text, an original short-text document was passed to pre-trained BERT. At each transformer layer a new word embedding was formed and saved. BERT's vector decoder was then used to convert hidden word vectors to candidate words, the top three candidate words at each encoder layer were kept. 
\newline
Each input word produced 48 candidate words, however many were duplicates. Examples of generated words per layer can be found in table \ref{tab:crime layers} and \ref{tab:politics layers} in the appendix. The generated words were sorted based on frequency, duplicate words from the original input were removed, as were stop-words, punctuation, and incomplete words. The generated words were then appended to the original document to create extended pseudo documents, the extended document was limited to 120 words in order to normalize each feature set. Further analysis can be found in table \ref{tab:data info} in the appendix.

\begin{Figure}
 \centering
 \includegraphics[width=\linewidth]{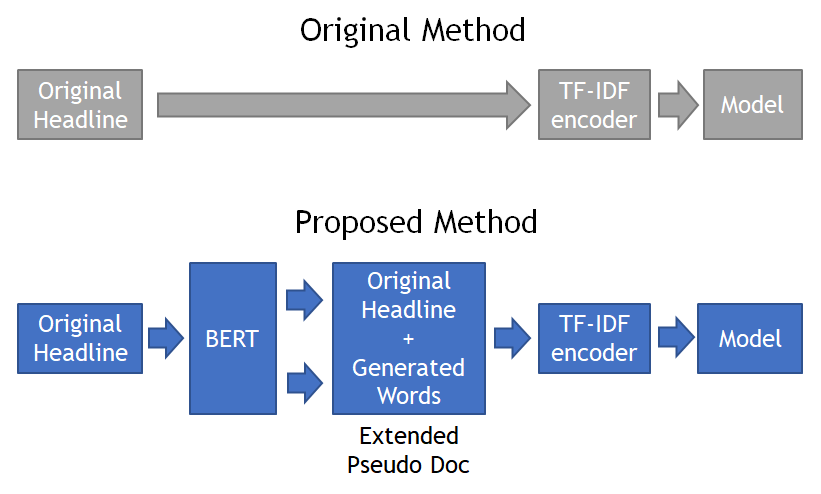}
 \captionof{figure}{The proposed method uses the BERT pre-trained word embedding model to generate new words which are appended to the orignal text creating extended pseudo documents.}
 \label{fig:Worldpost_vs_Crime}
\end{Figure}

\subsection{Topic Evaluation}
\label{topic relevance}
To test the proposed methods ability to generate unsupervised words, it was necessary to devise a method of measuring word relevance. Topic modeling was used based on the assumption that words found in the same topic are more relevant to one another then words from different topics \cite{steyvers2007probabilistic}. The complete 200k headline dataset \cite{dataset} was modeled using a Naïve Bayes Algorithm \cite{zhang2004optimality} to create a word-category co-occurrence model. The top 200 most relevant words were then found for each category and used to create the topic table \ref{tab:Top_word_table}. It was assumed that each category represented its own unique topic. \newline
The number of relevant output words as a function of the headline’s category label were measured, and can be found in figure \ref{fig:cat box plot}. The results demonstrate that the proposed method could correctly identify new words relevant to the input topic at a signal to noise ratio of 4 to 1.

\begin{Figure}
 \centering
 \includegraphics[width=200pt]{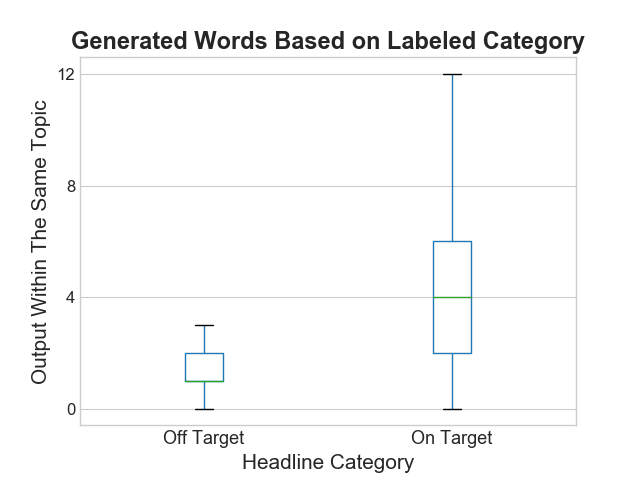}
 \captionof{figure}{The number of generated words within each topic was counted, topics which matched the original headline label were considered 'on target'. Results indicate that the unsupervised generation method produced far more words relating to the label category then to other topics. Tested on 7600 examples spanning 23 topics.}
 \label{fig:cat box plot}
\end{Figure}

\subsection{Binary and Multi-class Classification Experiments}

Three datasets were formed by taking equal length samples from each category label. The new datastes are \textit{‘Worldpost vs Crime’},  \textit{‘Politics vs Entertainment’}, and \textit{‘Sports vs Comedy’}, a fourth multiclass dataset was formed by combining the three above sets. \newline
For each example three feature options were created by extending every headline by 0, 15 and 120 words. 
Before every run, a test set was removed and held aside. The remaining data was sampled based on the desired training size. Each feature option was one-hot encoded using a unique tfidf-vectorizer \cite{scikit-learn} and used to train a random-forest classifier \cite{breiman2001random} with 300-estimators for binary predictions and 900-estimators for multiclass.
\newline
Random forest was chosen since it performs well on small datasets and is resistant to overfitting \cite{segal2004machine}. Each feature option was evaluated against its corresponding test set. 10 runs were completed for each dataset.

\section{Results and Analysis}
\subsection{Evaluating word relevance}
It is desirable to generate new words which are relevant to the target topics and increase predictive signal, while avoiding words which are irrelevant, add noise, and mislead predictions. \newline 
The strategy, described in section \ref{topic relevance}, was created to measure word relevance and quantify the unsupervised model performance. It can be seen from fig \ref{fig:cat box plot} and \ref{fig:Box_Plot} in the appendix that the proposed expansion method is effective at generating words which relate to topics of the input sentence, even from very little data. From the context of just a single word, the method can generate 3 new relevant words, and can generate as many as 10 new relevant words from sentences which contain 5 topic related words \ref{fig:Box_Plot}. While the method is susceptible to noise, producing on average 1 word related to each irrelevant topic, the number of correct predictions statistically exceed the noise. \newline
Furthermore, because the proposed method does not have any prior knowledge of its target topics, it remains completely domain agnostic, and can be applied generally for short text of any topic. 

\subsubsection{Binary Classification}
Comparing the performance of extended pseudo documents on three separate binary classification datasets shows significant improvement from baseline in the sparse data region of 100 to 1000 training examples.
\begin{itemize}
\item The ‘Worldpost vs Crime’ dataset showed the most improvement as seen in figure \ref{fig:Worldpost_vs_Crime}. Within the sparse data region the extended pseudo documents could achieve similar performance as original headlines with only half the data, and improve F1 score between 13.9\% and 1.7\% 
\item The ‘Comedy vs Sports’ dataset, seen in figure \ref{fig:Comedy_vs_Sports}, showed an average improvement of 2\% within the sparse region. 
\item The ‘Politics vs Entertainment’ dataset, figure \ref{fig:Politics_vs_Entertainment}, was unique. It is the only dataset for which a 15-word extended feature set surpassed the 120-words feature set. It demonstrates that the length of the extended pseudo documents can behave like a hyper parameter for certain datasets, and should be tuned according to the train-size.
\end{itemize}

\subsubsection{Multiclass Classification}
The Extended pseudo documents improved multiclass performance by 4.6\% on average, in the region of 100 to 3000 training examples, as seen in figure \ref{fig:6 class Multiclass}. The results indicate the effectiveness of the proposed method at suggesting relevant words within a narrow topic domain, even without any previous domain knowledge. \newline

In each instance it was found that the extended pseudo documents only improved performance on small training sizes. This demonstrates that while the extended pseudo docs are effective at generating artificial data, they also produce a lot of noise. Once the training size exceeds a certain threshold, it becomes no longer necessary to create additional data, and using extended documents simply adds noise to an otherwise well trained model.

\begin{Figure}
 \centering
 \includegraphics[width=\linewidth]{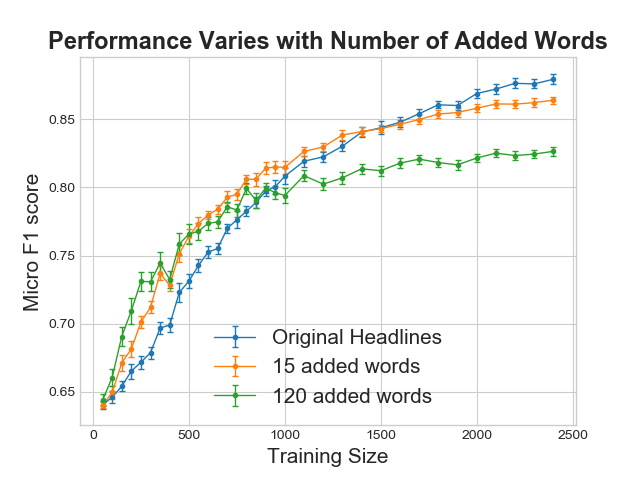}
 \captionof{figure}{Binary Classification of 'Politics' or 'Entertainment' demonstrates that the number of added words can behave like a hyper paremeter and should be tuned based on training size. Tested on 1000 examples with 10-fold cross validation}
 \label{fig:Politics_vs_Entertainment}
\end{Figure}

\begin{Figure}
 \centering
 \includegraphics[width=\linewidth]{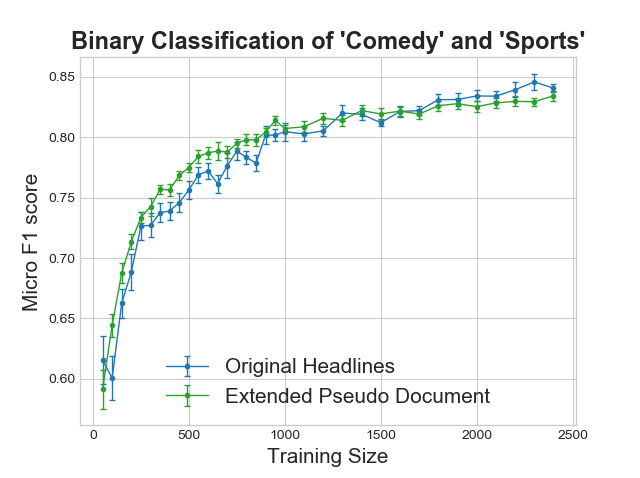}
 \captionof{figure}{Binary Classification of 'Politics' vs 'Sports' has less improvement compared to other datasets which indicates that the proposed method, while constructed to be domain agnostic, shows better performance towards certain topics. Tested on 1000 examples with 10-fold cross validation.}
 \label{fig:Comedy_vs_Sports}
\end{Figure}

\end{multicols}

\begin{Figure}
 \centering
 \includegraphics[width=\linewidth]{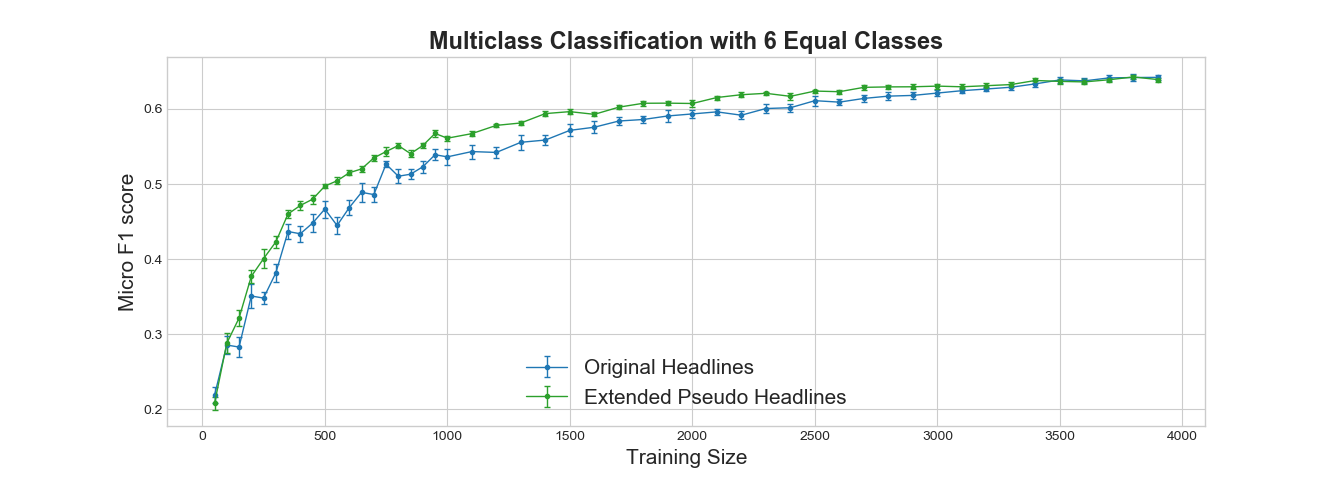}
 \captionof{figure}{Added Words improve Multiclass Classification between 1.5\% and 13\% in the range of 150 to 2000 training examples. Tests were conducted using equal size samples of Headlines categorized into 'World-Post', 'Crime', 'Politics', 'Entertainment', 'Sports' or 'Comedy'. A 900 Estimator Random Forest classifier was trained for each each data point, tested using 2000 examples, and averaged using 10-fold cross validation.}
 \label{fig:6 class Multiclass}
\end{Figure}

\begin{multicols}{2}

\section{Discussion}
Generating new words based solely on ultra small prompts of 10 words or fewer is a major challenge. A short sentence is often characterized by a just a single keyword, and modeling topics from such little data is difficult. Any method of keyword generation that overly relies on the individual words will lack context and fail to add new information, while attempting to freely form new words without any prior domain knowledge is uncertain and leads to misleading suggestions. \newline
This method attempts to find balance between synonym and free-form word generation, by constraining words to fit the original sentence while still allowing for word-word and word-sentence interactions to create novel outputs. \newline
The word vectors must move through the transformer layers together and therefore maintain the same token order and semantic meaning, however they also receive new input from the surrounding words at each layer. The result, as can be seen from table \ref{tab:crime layers} and \ref{tab:politics layers} in the appendix, is that the first few transformer layers are mostly synonyms of the input sentence since the word vectors have not been greatly modified. The central transformer layers are relevant and novel, since they are still slightly constrained but also have been greatly influenced by sentence context. And the final transformer layers are mostly non-sensical, since they have been completely altered from their original state and lost their ability to retrieve real words. \newline
This method is unique since it avoids needing a prior dataset by using the information found within the weights of a general language model. Word embedding models, and BERT in particular, contain vast amounts of information collected through the course of their training. BERT Base for instance, has 110 Million parameters and was trained on both Wikipedea Corpus and BooksCorpus \cite{devlin2018bert}, a combined collection of over 3 Billion words. The full potential of such vastly trained general language models is still unfolding. This paper demonstrates that by carefully prompting and analysing these models, it is possible to extract new information from them, and extend short-text analysis beyond the limitations posed by word count.


\end{multicols}


\newpage

\section{Appendix}
\subsection{Additional Tables and Figures}

\begin{Table}
    \centering \includegraphics[width=\linewidth]{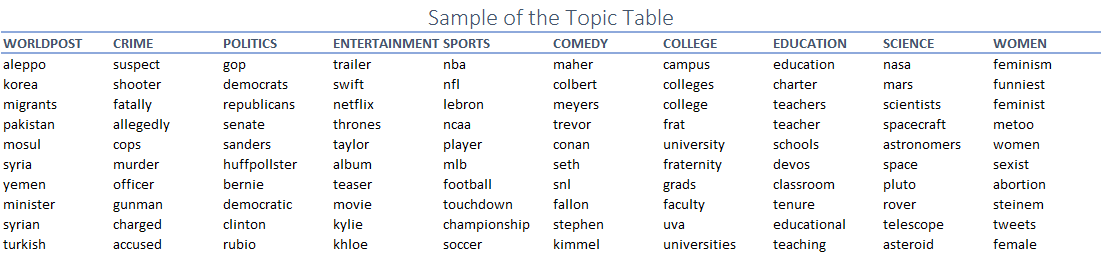}
    \captionof{figure}{A Topic table, created from the category labels of the complete headline dataset, can be used to measure the relevance of generated words.}
    \label{tab:Top_word_table}
\end{Table}

An original headline was analyzed by counting the number of words which related to each topic. The generated words were then analyzed in the same way. The change in word count between input topics and output topics was measured and plotted as seen in figure \ref{fig:Box_Plot}.  \newline
\begin{Figure}
 \centering
 \includegraphics[width=200pt]{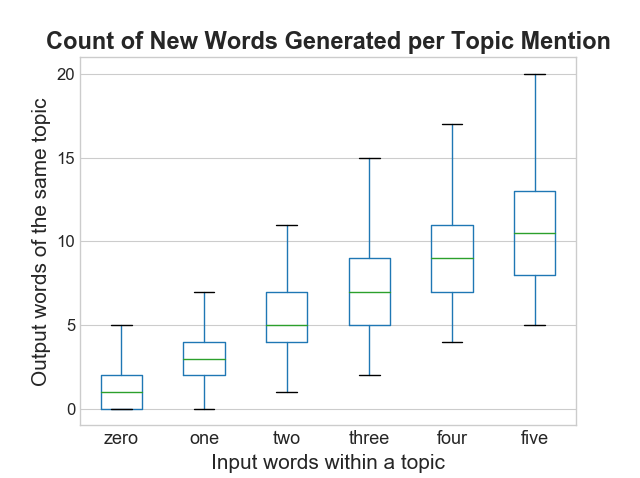}
 \captionof{figure}{Box plot of the number of generated words within a topic as a function of the number of input words within the same topic. Results indicate that additional related words can be generated by increasing the signal of the input prompt. Tested on 7600 examples spanning 23 topics.}
 \label{fig:Box_Plot}
\end{Figure}

\begin{Table}
    \centering \includegraphics[width=120pt]{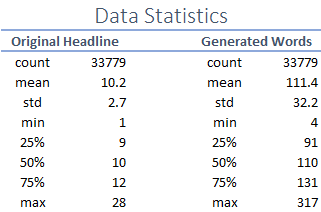}
    \captionof{figure}{Information regarding the original headlines, and generated words used to create extended pseudo headlines.}
    \label{tab:data info}
\end{Table}

\begin{Table}
    \centering \includegraphics[width=300pt]{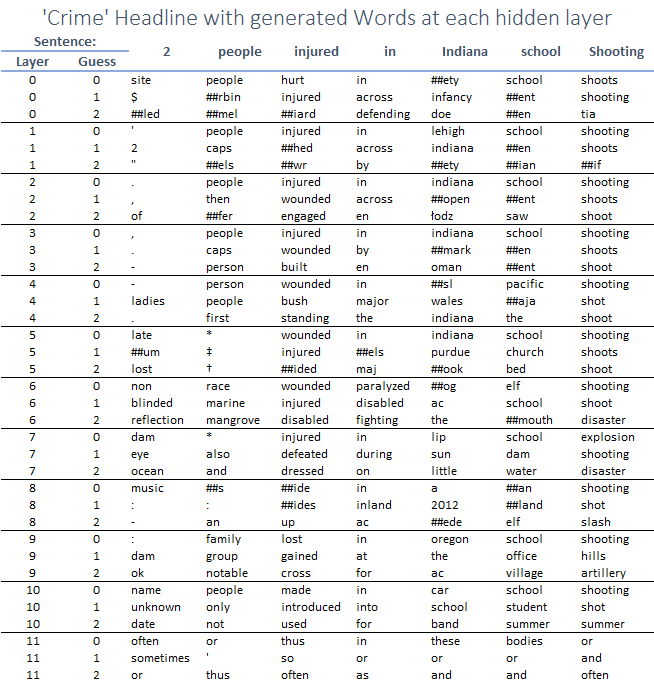}
    \captionof{figure}{Top 3 guesses for each token position at each later of a BERT pretrained embedding model. Given the input sentence '2 peoplpe injured in Indiana school shooting', the full list of generated words can be obtainedfrom the values in the table. }
    \label{tab:crime layers}
\end{Table}

\begin{Table}
    \centering \includegraphics[width=400pt]{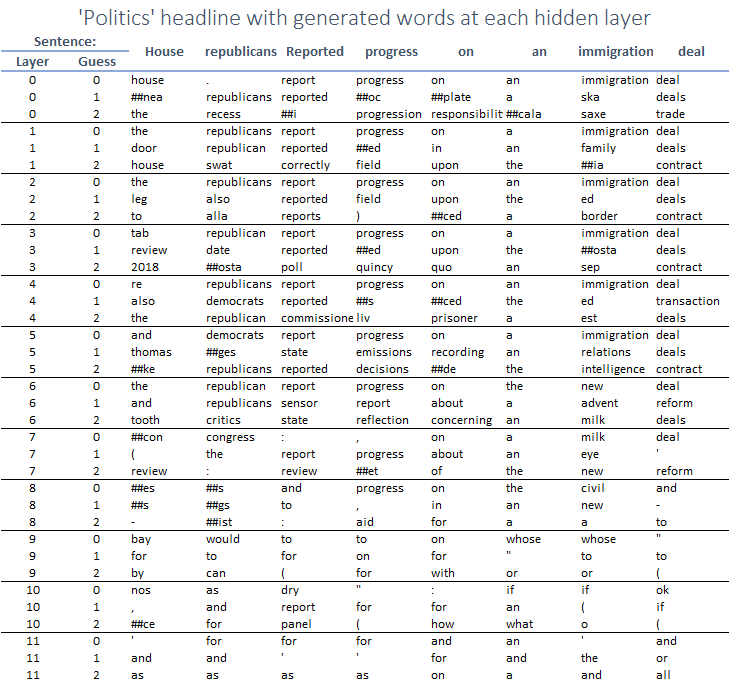}
    \captionof{figure}{Top 3 guesses for each token position at each later of a BERT pretrained embedding model.}
    \label{tab:politics layers}
\end{Table}

\end{document}